# Analysis of Semi-Supervised Learning with the Yarowsky Algorithm

**GholamReza Haffari** and **Anoop Sarkar**
School of Computing Science
Simon Fraser University
{ghaffar1,anoop}@cs.sfu.ca

## Abstract

The Yarowsky algorithm is a rule-based semi-supervised learning algorithm that has been successfully applied to some problems in computational linguistics. The algorithm was not mathematically well understood until (Abney 2004) which analyzed some specific variants of the algorithm, and also proposed some new algorithms for bootstrapping. In this paper, we extend Abney's work and show that some of his proposed algorithms actually optimize (an upper-bound on) an objective function based on a new definition of cross-entropy which is based on a particular instantiation of the Bregman distance between probability distributions. Moreover, we suggest some new algorithms for rule-based semi-supervised learning and show connections with harmonic functions and minimum multi-way cuts in graph-based semi-supervised learning.

## 1 Introduction

This paper addresses the semi-supervised learning problem where there is training data composed of some labeled as well as some unlabeled data points. The Yarowsky algorithm is a semi-supervised learning algorithm that has produced successful results in computational linguistics (Yarowsky 1995). It is an iterative bootstrapping algorithm where in each iteration a classifier is built based on the current set of labeled data, the learned classifier is used to produce label for unlabeled data, and a subset of newly labeled data contributes new features for the classifier trained for the next iteration. Despite success in various experimental studies, the Yarowsky algorithm was not mathematically well understood until (Abney 2004) which tried to show that the Yarowsky algorithm minimizes a reasonable objective function. In fact (Abney 2004) did not give an objective function for the original algorithm in (Yarowsky 1995) but introduced some variants of the original algorithm which were shown to optimize reasonable objective functions.

In this paper, we extend Abney's work and show that some of his proposed algorithms, which are very similar to the original Yarowsky algorithm, optimize (an upper-bound on) a reasonable objective function based on a particular instantiation of the Bregman distance between probability distributions and a new definition of cross-entropy. Our analysis and the contrast with the analysis in (Abney 2004) is given in Section 3. With our analysis, we hope to provide a general theoretical framework for the analysis of bootstrapping (or self-training) algorithms for semi-supervised learning with the use of Bregman distances. In Section 4 we show how we can view variants of some well known models for graph-based semi-supervised learning, such as the models described in (Zhu et. al. 2003, Blum and Chawla 2001), as particular instantiations of our general model. In addition, in Section 5 we formulate a new variant of the Yarowsky algorithm which is suggested by our analysis in Section 3.

Co-training (Blum and Mitchell 1998) is another algorithm for learning from labeled and unlabeled data which assumes that each data point can be described by two distinct *views*. Co-training learns two classifiers, one for each view, and the PAC-learning analysis by (Dasgupta et. al. 2001) shows that the classifier trained on one view has low generalization error if it agrees on unlabeled data with the classifier trained on the other view. The assumption under which this result holds is that each view is conditionally independent of the other view given the class label. However, (Abney 2002) shows that this assumption of view independence is remarkably powerful, and often violated in real data.

In contrast to co-training, the Yarowsky algorithm uses only a single classifer. (Abney 2004) shows that we can view the Yarowsky algorithm as minimizing the label disagreement between pairs of features in the classifier. Our motivation in this paper is to further study the Yarowsky algorithm, as it could be an attractive alternative to the co-training algorithm in many settings in which we wish to explore learning by bootstrapping.

## 2 The Modified Yarowsky Algorithm

The modified Yarowsky algorithm is introduced in (Abney 2004) and is shown in Algorithm 1. Throughout this paper, $L$ is the number of possible labels, $X$ is the set of labeled and unlabeled examples, $\Lambda^{(t)}$ is the set of labeled instances at time step $t$ and $Y^{(t)}$ is a function that provides the label for each instance, $V^{(t)}$ is the set of unlabeled examples at time $t$, and $\pi_x^{(t+1)}(j)$ is the score of the model in predicting label $j$ for the example $x$ in the time step $t+1$.

The modified Yarowsky algorithm differs from the original algorithm (Yarowsky 1995) as follows:

- Once an unlabeled example gets labeled, it stays labeled (its label may change after recomputing the labels)
- The labeling threshold is fixed to be $\frac{1}{L}$ (instead of an additional threshold parameter as used in the original Yarowsky algorithm).

---

**Algorithm 1** The modified Yarowsky algorithm

1: Given: $X, Y^{(0)}$
2: **for** $t \in \{0, 1, ...\}$ **do**
3: $\quad \Lambda^{(t)} = \{x \in X \mid Y^{(t)} \neq \bot\}$
4: $\quad$ Train classifier $\pi^{t+1}$ on $(\Lambda^{(t)}, Y^{(t)})$
5: $\quad$ **for** each instance $x \in X$ **do**
6: $\quad\quad \hat{y} := \arg\max_j \pi_x^{(t+1)}(j)$
7: $\quad\quad Y_x^{(t+1)} := \begin{cases} Y_x^{(0)} & \text{if } x \in \Lambda^{(0)} \\ \hat{y} & \text{if } x \in \Lambda^{(t)} \vee \pi_x^{(t+1)}(\hat{y}) > \frac{1}{L} \\ \bot & \text{otherwise} \end{cases}$
8: $\quad$ **end for**
9: $\quad$ **if** $Y^{(t+1)} = Y^{(t)}$ **then** Stop
10: **end for**

---

### 2.1 The Objective Function

Define $\phi_x(j)$ to be the probability that instance $x$ belong to the $j$th class. If a labeled instance $x$ belongs to the class $j$, the labeling distribution $\phi_x$ has all of its mass on the label $j$, and if $x$ is unlabeled the labeling distribution is uniform over all possible labels. The proposed objective function is the cross entropy between the prediction distribution of the model $\pi_x$ and the labeling distribution $\phi_x$ over *all* (labeled and unlabeled) instances:

$$l(\phi, \theta) \triangleq \sum_{x \in X} H(\phi_x \parallel \pi_x) \quad (1)$$

where $H(\phi_x \parallel \pi_x) = \sum_j \phi_x(j) \log \frac{1}{\pi_x(j)}$. Recall that

$$H(\phi_x \parallel \pi_x) = H(\phi_x) + D(\phi_x \parallel \pi_x) \quad (2)$$

where $H(\phi_x) = -\sum_j \phi_x(j) \log \phi_x(j)$ is the Shannon entropy, and $D(\phi_x \parallel \pi_x) = \sum_j \phi_x(j) \log \frac{\phi_x(j)}{\pi_x(j)}$ is the KL-divergence.

Consider the objective function (1). For labeled instances the entropy of the labeling distribution is zero, and the contribution to the objective function is $\sum_{x \in \Lambda} D(\phi_x \parallel \pi_x)$. This implies that the model parameters should be chosen so that the likelihood of the labeled data is maximized. For an unlabeled data point, the contribution is $H(\phi_x \parallel \pi_x) = H(\phi_x) + D(\phi_x \parallel \pi_x)$ which is minimized when the entropy goes to zero, i.e. it becomes labeled, and its assigned label agrees with the model prediction. Therefore minimizing the objective function forces unlabeled data to be labelled, and forces the model to maximize the likelihood of the (old and newly) labeled data.[1]

## 3 Specific Yarowsky algorithms

In this section we introduce some variants of Algorithm 1 defined in (Abney 2004) which use a specific base learner. We then provide our analysis of one of these variants.

### 3.1 The DL-0 algorithm

In the original Yarowsky algorithm (henceforth, the DL-0 algorithm), the base learner builds a decision list which consists of rules $f \to j$ showing that feature $f$ predicts label $j$ with the confidence score $\theta_{fj}$. The number of all possible features is $N$ and we assume that each instance $x$ has a subset $F_x$ of them. We can represent the relation between the features $F_x \in F$ and instances $x \in X$ as a bipartite graph as shown in Figure 1. Each example (in the right column) is connected to its features (in the left column). The labeled data have fixed labels, e.g. in Figure 1 one point is labeled '+' and one is labeled '−'. This representation is similar to the one used in (Corduneanu 2006).

---

[1] Eqn (1) is different from the objective function of conditional entropy regularization (Grandvalet and Bengio 2005): $\sum_{x \in \Lambda^{(0)}} H(\phi_x \parallel \pi_x) + \gamma \sum_{x \in V^{(0)}} H(\pi_x)$ where $\Lambda^{(0)}$ is the labeled data, $V^{(0)}$ is the unlabeled data, and $\gamma$ is used to control influence of unlabeled examples.

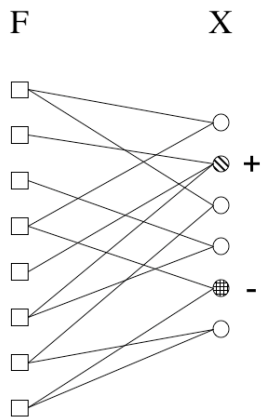

Figure 1: A bipartite graph representing the relationship between data points $X$ and features $F$. Each data point has been connected to its features by undirected edges, and some data points are initially labeled.

We further develop the relation between DL-0 and this bipartite graph representation in Section 4.

Among the rules matching the features of an instance $x$, the DL-0 algorithm selects the one which has the highest score and assigns the label predicted with that rule to $x$. In our terminology, this is equivalent to defining

$$\pi_x(j) \propto \max_{f \in F_x} \theta_{fj} \qquad (3)$$

and then assigning the label 'arg $\max_j \pi_x(j)$' to $x$. We assume the scores are non-negative and bounded, so they can be normalized for each instance to form a prediction distribution $\pi_x$.

The base learner in the DL-0 algorithm uses the smoothed precision to update $\theta_{fj}$. Define the precision $q_{fj}$ of a rule $f \to j$ to be

$$q_{fj} = \frac{|\Lambda_{fj}|}{|\Lambda_f|} \qquad (4)$$

where $\Lambda_{fj}$ is the set of all instances having the feature $f$ and the label $j$, and $\Lambda_f$ is the set of all instances having the feature $f$. Smoothed precision is defined by adding $\epsilon$ to the counts in the numerator:

$$\hat{q}_{fj} = \frac{|\Lambda_{fj}| + \epsilon}{|\Lambda_f| + L\epsilon} \qquad (5)$$

In the Yarowsky algorithm, a rule's score is taken to be its smoothed precision $\theta_{fj} = \hat{q}_{fj}$. Note that $\theta_{fj}$ are the model parameters which characterize the classifier in a space of classifiers $\Pi$. Both raw precision and smoothed precision can be viewed as a conditional probability distribution. We take the vector $\theta_f \equiv (\theta_{f1}, \dots, \theta_{fL})$ to have the properties of a conditional probability distribution $\theta_{fj} = p(j \mid f)$ in the rest of this paper.

It is not known whether the DL-0 algorithm, with the definition (3) for the prediction distribution and updating rule (5) for the parameters, minimizes the objective function $H$ defined in (1).

### 3.2 The DL-1 Algorithm

The DL-1 algorithm differs in two ways from the DL-0 algorithm. The first difference is in the definition of the prediction distribution. For an instance $x$ the prediction distribution is constructed as follows:

$$\pi_x(j) \propto \sum_{f \in F_x} \theta_{fj} \qquad (6)$$

The scores of each feature sum to one, hence $\pi_x(j) = \frac{1}{|F_x|} \sum_{f \in F_x} \theta_{fj}$. The definition (6) says that for an instance $x$ the prediction distribution is a mixture of its features' distributions.

Second, the DL-1 algorithm[2] uses the updating rule for parameters which differs slightly from smoothed precision (5):

$$\theta_{fj} = \frac{|\Lambda_{fj}| + \frac{1}{L}|V_f|}{|\Lambda_f| + |V_f|} \qquad (7)$$

where the smoothing constant is different for different features, in fact $\epsilon$ in (5) is replaced with $\epsilon_f = \frac{1}{L}|V_f|$.

#### 3.2.1 Analysis of the DL-1 Algorithm

We show that the DL-1 algorithm optimizes the upper-bound $K_{t^2}$ (which will be defined shortly) on the objective function (1) but based on a new definition of cross entropy. As we will see, the new definition of cross entropy is naturally derived based on a particular instantiation of the Bregman distance between two probability distributions.

Let $\psi$ be a strictly convex real-valued function, the Bregman distance $B_\psi(\mathbf{p}, \mathbf{q})$ between two discrete probability distributions $\mathbf{p}$ and $\mathbf{q}$ is defined as follows (Lafferty 1997):

$$\begin{aligned} B_\psi(\mathbf{p}, \mathbf{q}) &\triangleq \sum_i \psi(p_i) - \psi(q_i) - \psi'(q_i)(p_i - q_i) \\ &= \Psi(\mathbf{p}) - \Psi(\mathbf{q}) - \nabla\Psi(\mathbf{q}) \cdot (\mathbf{p} - \mathbf{q}) \end{aligned}$$

and the $\psi$-entropy of a discrete probability distribution $\mathbf{p}$ is defined as $H_\psi(\mathbf{p}) = -\sum_i \psi(p_i)$. When $\psi(t) = t^2$ we obtain the mean squared distance $B_{t^2}(\mathbf{p}, \mathbf{q}) = \sum_i (p_i - q_i)^2$, and the entropy as $H_{t^2}(\mathbf{p}) = -\sum_i p_i^2$. Moreover, if we let $\psi(t) = t \log t$, then $H_{t \log t}$ is the Shannon entropy and $B_{t\log t}(\mathbf{p}, \mathbf{q})$ is the KL-Divergence between these two probability distributions.

---

[2]In fact two variants of the DL-1 are introduced in (Abney 2004): (i) DL-1-R which uses the the raw precision (4) for updating the parameters, and (ii) DL-1-VS. In our discussion, we just analyze DL-1-VS and simply call it DL-1.

In the original formulation (2), cross entropy is equal to the Shannon entropy plus KL-Divergence. So we define the $\psi$-cross entropy between $\mathbf{p}$ and $\mathbf{q}$ to have the same relation:

$$H_\psi(\mathbf{p} \parallel \mathbf{q}) \triangleq H_\psi(\mathbf{p}) + B_\psi(\mathbf{p}, \mathbf{q})$$
$$= -\Psi(\mathbf{q}) - \nabla \Psi(\mathbf{q}) \cdot (\mathbf{p} - \mathbf{q})$$

In particular when $\psi(t) = t^2$, we will have $H_{t^2}(\mathbf{p} \parallel \mathbf{q}) = \sum_j q_j^2 - 2p_j q_j$. Let $l_{t^2}(\phi, \theta)$ refer to $l(\phi, \theta)$ when the new definition of cross entropy is used; then

$$l_{t^2}(\phi, \theta) = \sum_{x \in X} H_{t^2}(\phi_x \parallel \pi_x)$$
$$= \sum_{x \in X} \sum_j \pi_x^2(j) - 2\pi_x(j)\phi_x(j) \quad (8)$$

We will show that the DL-1 algorithm minimizes the following objective function

$$K_{t^2}(\phi, \theta) \triangleq \sum_{x \in X} \sum_{f \in F_x} \left[ H_{t^2}(\phi_x \parallel \theta_f) \right]$$

where $\frac{1}{m} K_{t^2}(\phi, \theta)$ is an upper-bound on $l_{t^2}(\phi, \theta)$ if we assume that instances have the same number of features, i.e. $|F_x|$ equals $m$ for each instance $x$. This assumption is also used in (Abney 2004) and is easy to satisfy when the instances do not have the same number of features by simply adding new features to the model.

**Lemma 1** $l_{t^2}(\phi, \theta)$ is upper-bounded by $\frac{1}{m} K_{t^2}(\phi, \theta)$.

**Proof**

$$\frac{1}{m} K_{t^2}(\phi, \theta) - l_{t^2}(\phi, \theta) =$$
$$\frac{1}{m^2} \sum_{x \in X} \sum_j \left[ m \sum_{f \in F_x} \theta_{fj}^2 - \left( \sum_{f \in F_x} \theta_{fj} \right)^2 \right] \geq 0$$

which is true for all integers $m \geq 1$ based on the fact that $|F_x| = m$. ∎

**Theorem 2** *The DL-1 algorithm minimizes the objective function $\frac{1}{m} K_{t^2}(\phi, \theta)$ which is an upper-bound for $l_{t^2}(\phi, \theta)$.*

**Proof** In each iteration of the DL-1 algorithm, we first update the parameters $\theta_{fj}$ and then recompute the labeling distribution $\phi_x$ for each example $x \notin \Lambda^0$.

Having the parameters $\theta_{fj}$ fixed, we show that the labeling distributions chosen by the algorithms reduce the contribution of each instance $x$ in $l_{t^2}(\phi, \theta)$ which is denoted by $l_{t^2}(\phi_x, \theta)$. There are three possible cases for the labeling of an instance $x \notin \Lambda^0$ in the previous and current iterations:

1) $x$ is unlabeled according to $\phi_x^{(t-1)}$ and $\phi_x^{(t)}$. It means both distributions are uniform and $\phi_x^{(t-1)} = \phi_x^{(t)}$, so $l_{t^2}(\phi_x, \theta)$ has not been changed after recomputing the labeling distribution.

2) $x$ is labeled according to $\phi^{(t-1)}$ and $\phi^{(t)}$. The label chosen by the DL-1 algorithms is $\arg\max_j \pi_x(j)$ which reduces $l_{t^2}(\phi_x, \theta)$ as much as possible. We view $l_{t^2}(\phi_x, \theta)$ as a function of $\phi_x$:

$$l_{t^2}(\phi_x, \theta) = \sum_{f \in F_x} \sum_j \theta_{fj}^2 - 2\phi_x(j)\theta_{fj}$$
$$= \text{Const} - 2|F_x| \sum_j \phi_x(j) \frac{\sum_{f \in F_x} \theta_{fj}}{|F_x|}$$
$$= \text{Const} - 2|F_x| \sum_j \phi_x(j)\pi_x(j)$$

3) $x$ is unlabeled according to $\phi^{(t-1)}$ but is labeled according to $\phi^{(t)}$. Before recomputing the labeling distribution $l_{t^2}(\phi_x, \theta) = \text{Const} - 2|F_x|\frac{1}{L}$ since $\phi_x^{(t-1)}$ is a uniform distribution. After recomputing the labeling distribution $l_{t^2}(\phi_x, \theta) = \text{Const} - 2|F_x| \max_j \pi_x(j)$. It is clear that $\max_j \pi_x(j) > \frac{1}{L}$ (because the instance becomes labeled), therefore the value of $l_{t^2}(\phi_x, \theta)$ has been reduced by labeling the instance.

Now assume that $\phi$ is fixed; the goal is to prove that when the algorithms update the the parameters $\theta_{fj}$, the objective function $l_{t^2}(\phi, \theta)$ is reduced. The Lagrangian for the constrained optimization problem is

$$\mathcal{L}(\theta, \lambda) = \sum_{x \in X} \Big( \sum_{f \in F_x} \sum_j \theta_{fj}^2 - 2\theta_{fj}\phi_x(j) \Big)$$
$$+ \sum_{f \in F} \lambda_f \Big( \sum_j \theta_{fj} - 1 \Big)$$

The optimality condition $\partial \mathcal{L}(\theta \lambda)/\partial \theta_{fk} = 0$ yields

$$\partial \mathcal{L}(\theta, \lambda)/\partial \theta_{fk} = \sum_{x \in X_f} 2\theta_{fk} - 2\phi_x(k) + \lambda_f = 0$$
$$\Rightarrow \theta_{fk} = \frac{|\Lambda_{fk}| + \frac{1}{L}|V_{fk}|}{|X_f|}$$

which is exactly the updating rule for the parameters in DL-1 algorithm. ∎

If we do not assume that instances have the same number of features (namely $m$), it is still true that the DL-1 algorithm minimizes $K_{t^2}(\phi, \theta)$. However it would not necessarily be the case that $l_{t^2}$ is upper-bounded by $\frac{1}{m} K_{t^2}$.

The analysis in (Abney 2004) provided an upper-bound on the objective function for DL-1, but based on the conventional definition of cross-entropy:

$$K(\phi, \theta) \triangleq \sum_{x \in X} \sum_{f \in F_x} \left[ H(\phi_x \parallel \theta_f) \right]$$

$$= \sum_{x \in X} \sum_{f \in F_x} \sum_j \phi_x(j) \log \frac{1}{\theta_{fj}}$$

This definition leads to an error in the proof provided in (Abney 2004) which attempts to show the relation between minimizing $K(\phi, \theta)$ and the label chosen by the DL-1 algorithm. The DL-1 algorithm chooses the label $\arg\max_j \pi_x(j) = \arg\max_j \sum_j \theta_{fj}$. However, $K$ is minimized as a function of $\phi$ (which is done per instance $x$) by selecting the label $\arg\min_j \sum_{f \in F_x} \log \frac{1}{\theta_{fj}}$. In general, $\arg\min_j \sum_{f \in F_x} \log \frac{1}{\theta_{fj}} \neq \arg\max_j \sum_j \theta_{fj}$. Our main result in this section, using the framework of (Abney 2004), provides an objective function $K_{t^2}(\phi, \theta)$ that, when minimized, provides the same labeling and parameter updates as the DL-1 algorithm.

## 4 Extensions of the DL-1 Algorithm

In this section we generalize the ideas in the previous section to present a general formulation of the graph-based semi-supervised learning. Consider Figure 1 which represents the relationship between instances and features by a bipartite graph: Each feature induces a *bias* to enforce the label similarity of its neighboring instances. In other words, the label of data points adjacent to a common feature should be similar. By noticing that data points may have overlapping features, this induces constraints among the labels of data points.

For the set of instances $X_f$ adjacent to a feature $f$, we assign a probability distribution (over labels) to $f$ and require the label of all of its neighbors to have a small distance to it by minimizing $\sum_{x \in X_f} B_\psi(\theta_f, \phi_x)$. The overall objective function for minimization is:

$$\sum_{f \in F} \sum_{x \in X_f} B_\psi(\theta_f, \phi_x) \quad (9)$$

The above objective function is simply the sum over the distances of probability distributions of nodes connected by edges in the graph. We will show that some well-known models like (Zhu et. al. 2003) can be seen as instantiations of this general model. In particular when $\psi(t) = t^2$, the objective function becomes:

$$\sum_{f \in F} \sum_{x \in X_f} \sum_{j=1}^L (\theta_{fj} - \phi_x(j))^2 \quad (10)$$

with the constraint that the labeling distribution of initially labeled points is fixed. It is not hard to see that this is a convex optimization program and at the optimum, the labeling distribution of every node in the graph should be the average of its neighbors. In the case of binary classification a real valued function is defined on the graph nodes, where the value of each node shows the probability of belonging to one of the output labels. In this setting the optimum function is *harmonic*, which is the mean of the Gaussian random field defined by the graph and can be computed by label propagation by performing a random walk on the graph (Zhu et. al. 2003). Here the vertices of the underlying graph consists of both instances and features, in contrast in (Zhu et. al. 2003) the vertices of the underlying graph are instances.

Suppose we want the labeling distribution of each data point to have a small entropy $-\sum_k \phi_x^2(k)$. We add the penalty term $-\sum_{x \in X} |F_x| \sum_j \phi_x^2(j)$ to the objective function (10) where the weights $|F_x|$ show the emphasis on different instances, and as the result we get the final objective function (8). The optimization problem is not convex, and the way that DL-1 algorithm propagates labels to reach a stationary solution is important. In addition to data points, imagine we further would like each feature to have a small entropy over its soft label. We can add another penalty term $-\sum_{f \in F}(|X_f| \sum_j \phi_x^2(j))$ to the objective function (the weights show the emphasis on different features), and with some simplification we get:

$$-2 \sum_{f \in F} \sum_{x \in X_f} \sum_{j=1}^L \phi_x(j) \times \theta_{fj} \quad (11)$$

In order to optimize the above objective function consider Algorithm 2 which is the modified Yarowsky algorithm specialized for this case. The function **majority**$(S)$ returns a probability distribution over the labels. If all labels are supported by equal number of nodes then the resulting probability distribution is uniform, otherwise the label which is supported by the majority of nodes in $S$ gets the total probability mass (if more than one label gets the highest number of supporters, one of them is chosen arbitrarily). Without loss of generality, we can ignore votes of the nodes in $S$ having the uniform distribution over labels. Note that in Algorithm 2 once an unlabeled node is labeled in some iteration, it retains that label even if the majority function returns the uniform distribution in a future iteration.

It is easy to prove that the objective function (11) can be minimized, as a function of $\theta_f$, by putting all of the mass of its probability distribution on the label $k$ which has the largest contribution $\sum_{x \in X_f} \theta_{xk}$. The largest value for $\sum_{x \in X_f} \theta_{xk}$ corresponds to the label which is the majority label in the neighbors of $f$. The same argument works for the second phase of the iterations where we update the label of instances. Moreover it can be shown that Algorithm 2 converges in polynomial time.

**Algorithm 2** Majority-Majority
1: **repeat**
2:   **for** $f \in F$ **do**
3:     $p \leftarrow \textbf{majority}(X_f)$
4:     **if** $p$ is **not** a uniform distribution **then**
5:       $\theta_f \leftarrow p$
6:   **end for**
7:   **for** $x \in X$ **do**
8:     $q \leftarrow \textbf{majority}(F_x)$
9:     **if** $q$ is **not** a uniform distribution **then**
10:      $\phi_x \leftarrow q$
11:  **end for**
12: **until** the labels do not change

**Theorem 3** *Algorithm 2 stops after at most $\mathcal{O}(|F|^2|X|^2)$ iterations.*

**Proof** In Appendix A.

Define a new operator **average**$(S)$ that returns a probability distribution which is the (normalized) sum of probability distributions associated with nodes in the set $S$. Replacing **majority** in line 3 of Algorithm 2 with **average** gives the DL-1 algorithm, and replacing **majority** in lines 3 *and* 8 with **average** gives the harmonic function based method.

Suppose we arbitrary break the ties and reach a stationary point where all nodes of the bipartite graph are labeled. Put all nodes from class 1 in one set, all nodes from class 2 in another set, and so on. Picking a node and changing its set will increase the cost of the $L$-way cut in the graph. In fact the algorithm greedily partitions the nodes into $L$ sets and reaches a (local) minimum $L$-way cut that agrees with the labeled instances. In the binary case, it is similar to semi-supervised learning using mincut (Blum and Chawla 2001) with the difference that here both instances and features are nodes of the underlying graph, but in (Blum and Chawla 2001) only the instances are the vertices of the underlying graph.

Consider minimizing the objective function in (9) for the general class of the Bregman distances with the constraint that the labeling distributions of the initially labeled data are fixed. The minimization is done on the distributions associated with features and unlabeled data. We may add more constraints and require the probability distributions belong to specific classes of models, e.g. $\theta_f \in \mathcal{M}_\mathcal{F}$ where $\mathcal{M}_\mathcal{F}$ is a particular class of probability distributions. For the case where $\mathcal{M}_\mathcal{F}$ is unrestricted, the optimal probability distribu-

tions must satisfy the following optimality conditions

$$\forall f \in F : \theta_f^{opt} \propto \nabla \Psi^* \left( \frac{1}{|X_f|} \sum_{x \in X_f} \nabla \Psi(\phi_x^{opt}) \right)$$

where $\Psi^*$ is the convex conjugate of the convex function $\Psi$, and

$$\forall x \in V : \sum_{f \in F_x} \left( \theta_{fj}^{opt} - \phi_x^{opt}(j) \right) \psi''(\phi_x^{opt}(j)) = c_x$$

$\forall j \in \{1,..,L\}$, where $c_x$ is a constant associated with instance $x$.

## 5    The DL-2 Algorithms

In this section we propose and analyze a new family of specific Yarowsky algorithms called the DL-2 family of algorithms. The algorithms in this family are named DL-2-S and DL-2-ML. The prediction distribution for the DL-2 family of algorithms is constructed by:

$$\pi_x(j) \propto \prod_{f \in F_x} \theta_{fj} \qquad (12)$$

with the normalization constant $Z_x = \sum_k \prod_{f \in F_x} \theta_{fk}$.

The intuition behind the prediction distribution (12) is as follows. Associated with each feature $f$, consider a biased die with $L$ faces and the parameter $\theta_{fj}$ for the $j$th face coming up. For an instance $x$, we toss all of the dice associated with features $f \in F_x$; what is the probability of seeing face $j$ in each die given we already know that all dice should show the same face? The answer is the expression given in (12). This way of combining predictions is similar to the Product of Experts (PoE) framework (Hinton 1999). In PoE, prediction distributions of simple models, or "experts", are multiplied together and then normalized to produce a sharp high dimensional prediction distribution. While equation (12) is similar to PoE, our parameter updates are very different, and we use it in order to provide a novel semi-supervised algorithm.

The base learner in DL-2-S is very similar to the base learner in DL-0 in that it uses the smoothed precision to update the parameters:

$$\theta_{fj} = \frac{|\Lambda_{fj}| + \frac{1}{L}(|V_f| + \delta|X_f|)}{|\Lambda_f| + |V_f| + \delta|X_f|} \qquad (13)$$

where $\delta$ is a given smoothness parameter. The smoothing constant is different for different features, in fact $\epsilon_f = \frac{1}{L}(|V_f| + \delta|X_f|)$.

In DL-2-ML, instead of learning a new classifier from scratch, the classifier in the previous iteration is improved to yield the new one. Updating the parameters is done in the direction of minimizing $H$. The only algorithm in this family which minimizes $H$ given in (1) is DL-2-ML.[3]

## 5.1 Analysis of DL-2-S Algorithm

We prove that DL-2-S minimizes the following objective function which is an upper-bound on $H$:

$$K_\delta \triangleq \sum_{x \in X} \sum_{f \in F_x} \left[ H(\phi_x \parallel \theta_f) + \delta \cdot H(u \parallel \theta_f) \right] \quad (14)$$

where $u$ is the uniform distribution over the labels. We start the analysis by proving the following lemma.

**Lemma 4** $H$ is upper-bounded by $K_\delta$ for all $\delta \geq 0$.

**Proof**

$$\prod_{f \in F_x} 1 = \prod_{f \in F_x} \sum_j \theta_{fj} =$$

$$= \underbrace{\sum_j \Big( \prod_{f \in F_x} \theta_{fj} \Big)}_{Z_x} + \text{some positive terms}$$

hence $\forall x, Z_x \leq 1$. Now we prove the lemma:

$$H = \sum_x H(\phi_x \parallel \pi_x)$$

$$= \sum_x \sum_j \phi_x(j) \Big( \log Z_x - \log \prod_{f \in F_x} \theta_{fj} \Big)$$

$$= \sum_x \Big( \log Z_x - \sum_f \sum_j \phi_x(j) \log \theta_{fj} \Big)$$

$$= \sum_x \Big( \log Z_x + \sum_f H(\phi_x \parallel \theta_f) \Big)$$

$$\leq \sum_x \sum_f H(\phi_x \parallel \theta_f)$$

The inequality holds because $\log a \leq 0$ for $a \leq 1$.

Since $\delta \geq 0$ and cross-entropy is always a non-negative function, it follows that $H \leq K_\delta$. ∎

The following lemma is used to prove the main theorem of this section.

**Lemma 5** *If the labeling distribution $\phi$ is fixed, then updating the parameters $\theta_{fj}$ based on the smoothed precision (13) minimizes $K_\delta$.*

**Proof** We should minimize $K_\delta$ subject to the constraints $\forall f, \sum_j \theta_{fj} = 1$. The Lagrangian associated

---

[3]Using a log-linear formulation of $\pi_x$ we can find a simple formula for the gradient of $H$ which can then be used by DL-2-ML to minimize $H$. However, due to lack of space, we omit further discussion of DL-2-ML from this paper.

with the optimization problem is

$$\mathcal{L}(\theta, \lambda) = \sum_{x \in X} \sum_{f \in F_x} \Big( H(\phi_x \parallel \theta_f) + \delta \cdot H(u \parallel \theta_f) \Big)$$

$$+ \sum_f \lambda_f \Big( \sum_j \theta_{fj} - 1 \Big)$$

$$= \sum_{x \in X} \sum_{f \in F_x} \Big( \sum_j \phi_x(j) \log \frac{1}{\theta_{fj}} + \delta \sum_j \frac{1}{L} \log \frac{1}{\theta_{fj}} \Big)$$

$$+ \sum_f \lambda_f \Big( \sum_j \theta_{fj} - 1 \Big)$$

The optimality condition is $\partial \mathcal{L}(\theta, \lambda) / \partial \theta_{fk} = 0$, which yields

$$\partial \mathcal{L}(\theta, \lambda) / \partial \theta_{fk} = - \sum_{x \in \Lambda_f} \Big[ \frac{\phi_x(k)}{\theta_{fk}} + \frac{\delta}{L \theta_{fk}} \Big]$$

$$- \sum_{x \in V_f} \Big[ \frac{1}{L \theta_{fk}} + \frac{\delta}{L \theta_{fk}} \Big] + \lambda_f = 0$$

hence the smoothed precision (13) updating rule is derived. ∎

**Theorem 6** *DL-2-S minimizes the objective function $K_\delta$ which is an upper-bound on $H$ for every $\delta \geq 0$.*

**Proof** The proof is similar to the proof of Theorem 2. In Lemma 4, it was proved that $K_\delta$ is an upper-bound on $H$ for every $\delta \geq 0$. In each iteration of the DL-2-S algorithm, we first update the parameters $\theta_{fj}$ and then recompute the labeling distribution $\phi_x$ for each example $x \notin \Lambda^0$. We show that each of these two phases reduces the objective function.

Having the parameters $\theta_{fj}$ fixed, the labeling distributions chosen by the algorithms reduces $K_\delta(x)$ for each instance $x$. If $x$ is labeled according to $\phi^{(t-1)}$ and $\phi^{(t)}$, the label chosen by the algorithms is

$$\arg\max_j \pi_x(j) = \arg\max_j \sum_f \log \theta_{fj} \quad (15)$$

Now look at $K_\delta(x)$ as a function of $\phi_x$:

$$K_\delta(x) = \sum_{f \in F_x} H(\phi_x \parallel \theta_f) + \delta \cdot H(u \parallel \theta_f)$$

$$= \sum_j \phi_x(j) \sum_{f \in F_x} \log \frac{1}{\theta_{fj}} + \text{Const}$$

It is clear that the labeling distribution chosen by the algorithms in (15) reduces $K_\delta(x)$ as much as possible (the other two cases where changing the label distribution reduces $K_\delta(x)$ is easy to verify).

Now assume that $\phi$ is fixed. In Lemma 5 it is proved that the DL-2-S algorithm updates the parameters in such a way that $K_\delta$ is minimized, which concludes the proof. ∎

As a special case if we set $\delta = 0$, the algorithm minimizes the following objective function

$$\sum_{x \in X} \sum_{f \in F_x} H(\phi_x \parallel \theta_f)$$

based on the following update rule for the parameters

$$\theta_{fj} = \frac{|\Lambda_{fj}| + \frac{1}{L}|V_f|}{|\Lambda_f| + |V_f|}$$

When $\delta = 0$ we obtain a direct relationship with the parameter update rule for the DL-1 algorithm in (7).

## 6 Conclusion

We analyzed several algorithms for rule-based semi-supervised learning. We showed that the DL-1 algorithm, which is a variant of the Yarowsky algorithm proposed in (Abney 2004), minimizes an upper-bound on a new definition of cross entropy based on a specific instantiation of the Bregman distance. The use of Bregman distances provide us with a general framework which we used in order to show strong connections to the graph-based semi-supervised learning algorithms of (Zhu et. al. 2003, Blum and Chawla 2001). By extending the DL-1 algorithm, we provide a simple polynomial-time algorithm for label propagation in graph-based semi-supervised learning. We also introduced a new class of algorithms for semi-supervised learning, called the DL-2 algorithms, which is shown to optimize a reasonable objective function.

### Acknowledgements

This research was partially supported by NSERC, Canada (RGPIN: 264905). We would like to thank Vahab S. Mirrokni, Valentine Kabanets, Oliver Schulte and the anonymous reviewers for their helpful comments.

## Appendix A. Convergence of Algorithm 2

**Lemma 7** *Suppose in some iteration in Algorithm 2, $r$ nodes in the right and $l$ nodes in the left columns are labeled in the bipartite graph (ignore the other nodes of the graph). The recomputation of the labels for these nodes (the main loop of the algorithm) terminates in no more than $l \times r$ iterations.*

**Proof** Suppose there are only two labels: $+1$ and $-1$. Put all $+1$ nodes in the set $A$ and all $-1$ nodes in the set $B$. Consider recomputing the label of a node $v$ in the graph: it is given the label which majority of its neighbors have. It means that if majority of its neighbors are in the set $A$ (or $B$) then this node is also put in the same set. After recomputing the label of a node $v$, if it is transferred from one set to another then the number of edges between sets $A$ and $B$, i.e. the *cut* size, is reduced by at least one. Now consider recomputing the labels of all nodes in the right (left) column of the graph in parallel: it reduces the cut size because the right (left) nodes do not have any edge among them. Each iteration, if a different label is produced for at least one node, causes the cut size to be reduced by at least one. The maximum number of edges between the right and left nodes is $r \times l$, and in the worst case the cut size reaches zero after $l \times r$ iterations or the algorithm terminates before this point. ∎

**Proof of Theorem 3** Consider the situation in which $r$ nodes in the right and $l$ nodes in the left are labeled. Due to Lemma 7, at the worst case after $l \times r$ iterations their labels are stabilized. However, the addition of even one node from the unlabeled nodes to this set of currently labeled data (before label stabilization) can stop the convergence of the algorithm, and cause it to continue for more iterations. Consider the worst case where only one node is added in the $(l \times r)$th iteration: it may be added to the left nodes or to the right nodes. So the number of iterations needed for the convergence of the next round of the algorithm is upper-bounded by $l \times (r+1) + (l+1) \times r$. Continuing this reasoning, it can be seen that the number of iterations needed for the convergence of the algorithm is at most $\sum_{i=1}^{|F|} \sum_{j=1}^{|X|} i \times j = \mathcal{O}(|F|^2|X|^2)$ ∎